\documentclass{article}

\usepackage{PRIMEarxiv}
\usepackage{cite}
\usepackage{amsmath,amssymb,amsfonts}
\usepackage{algorithmic}
\usepackage{textcomp}
\usepackage{xcolor}
\usepackage{subcaption}
\usepackage{balance}
\usepackage{todonotes}
\usepackage{multirow}%
\usepackage{booktabs}
\usepackage{comment}
\usepackage[utf8]{inputenc} 
\usepackage[T1]{fontenc}    
\usepackage{hyperref}       
\usepackage{url}            
\usepackage{booktabs}       
\usepackage{amsfonts}       
\usepackage{nicefrac}       
\usepackage{microtype}      
\usepackage{lipsum}
\usepackage{fancyhdr}       
\usepackage{graphicx}       
\graphicspath{{media/}}     
\def\BibTeX{{\rm B\kern-.05em{\sc i\kern-.025em b}\kern-.08em
    T\kern-.1667em\lower.7ex\hbox{E}\kern-.125emX}}
\title{A New AFIB Detection Method with Deep Neural Networks}

\author{
  Sara Gjorgjieva$^{1}$ $^{2}$ \href{https://orcid.org/0009-0009-2014-4515}
    {\raisebox{0.8ex}{\scriptsize[0009-0009-2014-4515]}}, 
  Ana Angjelevska$^{1}$ $^{2}$, 
  Dimitar Trajanov$^{2}$
    \href{https://orcid.org/0000-0002-3105-6010}
    {\raisebox{0.8ex}{\scriptsize[0000-0002-3105-6010]}},\\
  Marjan Gusev$^{1}$ $^{2}$ 
    \href{https://orcid.org/0000-0003-0351-9783}
    {\raisebox{0.8ex}{\scriptsize[0000-0003-0351-9783]}}\\
  \\
  $^{1}$Innovation Dooel, Skopje, North Macedonia \\
  $^{2}$Ss. Cyril and Methodius University in Skopje, Faculty of Computer Science and Engineering, North Macedonia \\
  \\
  \texttt{\{sara.gjorgjieva, ana.angjelevska\}@innovation.com.mk} \\
  \texttt{\{dimitar.trajanov, marjan.gushev\}@finki.ukim.mk}
}

\begin{document}
\maketitle

\vspace{-2em}
\begin{center}
\small
\textit{Published in 16th International Conference, ICT Innovations 2024 Web Proceedings},
pp.~140--153. Skopje: ICT-ACT, \\ September 2024.
\end{center}
\vspace{2em}

\begin{abstract}
Early detection of atrial fibrillation (AFIB), characterized by irregular heart rhythms, can potentially save thousands of lives annually  \cite{sankari2011heartsaver}.  Numerous studies have attempted to address this challenge using various algorithms and methodologies. This paper introduces a novel and robust method that enhances the precision of AFIB detection.
%
Neural Networks (NN) are widely recognized for their ability to identify complex electrocardiogram (ECG) patterns, facilitating the accurate detection of irregular heart rhythms. This study explores multiple conventional machine learning algorithms alongside neural network approaches to develop a robust AFIB detection system. To evaluate the effectiveness and robustness of the proposed method, the MIT BIH arrhythmia database (MITDB) and the Long-term Atrial Fibrillation Database (LTAFDB), both available on the Physionet site.
Our approach presents a novel feature engineering technique, employing specific features such as a time series of the differences in successive heart rates (dBPM) rather than traditional inter-beat intervals. The model analyzes inputs labeled into five distinct classes, using only clean intervals identified through a new labeling method for training while treating others as outliers. This strategy ensures a more accurate and reliable AFIB detection process.
%
The best-performing model, a Deep Neural Network (DNN), achieved an F1 score of 95.18\% for binary atrial fibrillation classification on a benchmark dataset distinct from the training set. Models trained using our approach have demonstrated superior performance than previously published models.
\end{abstract}

\keywords{atrial fibrillation \and arrhythmia \and deep learning \and ECG}

\section{Introduction}

Atrial fibrillation (AFIB) is a type of cardiac arrhythmia that occurs in the heart's atria due to disrupting the normal cycle of the heart's electrical impulses. It typically arises from increased automatism in a group of cells within the atria, causing their contractions to exceed the regular rate of impulses per minute and resulting in complete electrical disorganization within the atria. These abnormal impulses override the heart's natural pacemaker, disrupting its ability to maintain a regular rhythm and leading to an irregular heart rate \cite{afib}.



AFIB prevents the heart muscle from relaxing properly between contractions, reducing its efficiency and overall performance. The condition manifests as a highly irregular pulse rate. It can lead to numerous health issues, including rapid heartbeat, weakness, and dizziness, leading to blood clots, stroke, heart failure, and other heart-related complications \cite{afib}.

Detecting AFIB requires the ability to identify specific ECG characteristics. AFIB is an irregular rhythm of heartbeats that seems normal in width but varies in height and spacing without preceding polarization P-waves.
Most state-of-the-art studies focus on detecting an irregular rhythm in the ECG without considering the absence of P-waves, which are very difficult to detect due to their very low amplitude. An irregular rhythm may have patterns such as those classified with Bigeminy (B) and Trigeminy (T), while AFIB is an irregular rhythm without patterns.

State-of-the-art analysis shows that most Neural Network approaches use RR intervals as the time between consecutive heartbeats. In this paper, we use a strategy based on differences in heart rates between consecutive heartbeats as a novel feature proven to outperform other approaches. We used multiple techniques to achieve a robust real-time medical device application, including cross-validation and an additional validation dataset to prevent overfitting. For this purpose, we used the standard {\em MIT-BIH arrhythmia ECG benchmark database (MITDB)} \cite{moody2001impact} and the {\em Long-Term Atrial Fibrillation ECG Database (LTAFDB)} \cite{petrutiu2007abrupt} available on the Physionet site \cite{gusev2019performance}.

The paper follows the next structure. Related work in Section~\ref{sec-related} analyses the similar approaches. Section~\ref{sec-methods} presents the methods in this research, explaining the datasets, feature engineering, machine learning models, and evaluation methodology. Section~\ref{sec-results} addresses the performance evaluation from the experiments. Discussion in Section~\ref{sec-discussion} compares the achieved results with the related work.  
Section~\ref{sec-conclusion} addresses the conclusion and gives directions for future work.

%
%

\section{Related Work}
\label{sec-related}

Search for an effective algorithm to detect AFIB results in numerous approaches. Our analysis shows the application of different signal processing and Machine Learning (ML) models, including NNs. Evaluation methods differ from using proprietary datasets via standards, such as EC57 \cite{EC57} or EIC 60601-2-47 \cite{60601} to derive meaningful evaluation metrics. 

Authors who use the same dataset for training and testing reveal high results, although these models are not robust enough. Applying the same model to a different patient performs very poorly. An example includes \cite{martis2013application} declaring an F1 score of 98.97\% with their NN model. 
A model created by combining a recurrent NN and long-short-term memory (LSTM) \cite{faust2018automated} was trained and tested on the same MITDB dataset with a 10-fold cross-validation to reach an F1 score of 98.35\%. 

To overcome the robustness and achieve applicability, the researchers used the patient-wise method, with totally different datasets for the training and testing.
For example, \cite{wang2020atrial} uses different combinations of 
convolutional NNs and long-short-term memory (LSTM) frameworks. Although using a 70/30 splitting criteria on the same MITDB dataset, this approach reveals an F1 score of 82\%.
Recurrent NNs for detecting AFIB \cite{teijeiro2017arrhythmia} on the same dataset (MITDB) obtain an F1 score of 83.00\%.
A distributed deep NNs \cite{cai2020accurate} was used on a proprietary dataset collecting records from multiple hospitals and wearable ECG devices to detect AFIB with a 5-fold cross-validation, achieving an F1 score of 90.70\%.

Our approach belongs to the class of patient-wise data split for training and testing datasets. Contrary to other NN solutions, our approach uses features based on differences in rates from consecutive heartbeats, while the others use intervals between consecutive heartbeats.

We plan three experiments: the first using the same dataset for training and testing, the second using separate ECG benchmark tests, and the third dividing the MITDB into two equal datasets. Many other papers typically use cross-validation splits.


%
%
\section{Methods}
\label{sec-methods}

%

%
\subsection{Datasets}

Databases in this research are fundamental in studying heart diseases and are widely referenced in the scientific community, publicly available from the Physionet website \cite{PhysioNetDatabases}. 

\textbf{MITDB ECG database} \cite{moody2001impact} contains 30-minute ECG recordings from 48 patients, totaling 109,483 annotated heartbeats. This research excludes four records from paced patients (102, 104, 107, and 217). Consequently, the total number of annotated heartbeats analyzed is 97,924. It is a golden benchmark for evaluating newly developed algorithms and ML models, and in this research, training the new model uses MITDB. Only 10\% of records consist of AFIB rhythm episodes.

\textbf{LTAFDB database }\cite{petrutiu2007abrupt} consists of 24 to 25-hour ECG recordings from 83 individual patients diagnosed with AFIB, with over half of the records with AFIB rhythm. This database contains 8,903,169 annotated heartbeats, excluding two paced patients (6 and 113); we excluded them, totaling with 
8,682,368, but because it is enormous compared to the other dataset, we used a shortened version where we included 943,953 annotated heartbeats.


The records from both databases contain predominantly Normal Sinus Rhythm (NSR), including episodes of Bigeminy (B), Trigeminy (T), Bradycardia (SBR), Tachycardia (SVT or SVTA), and a small portion of other atrial and ventricular arrhythmia. This research excludes episodes of Ventricular Fibrillation (VFIB) and Ventricular Flutter (VFL) from the analysis.
Both databases are imbalanced, which is a common challenge in medical data. Addressing this imbalance will be crucial in training and evaluating our models effectively for atrial fibrillation detection.

%
\subsection{Feature Engineering}\label{subsec3}

Feature engineering is crucial in developing machine learning models, especially for detecting complex conditions such as atrial fibrillation (AFIB) from heart rate data. Feature engineering aims to extract features from raw data that can effectively capture the underlying patterns and differences between normal heart rhythms and AFIB episodes. For AFIB detection, various features derived from electrocardiogram (ECG) data enhance the model's ability to distinguish between different cardiac rhythms.

This research focuses on features derived from the intervals between heartbeats to develop robust machine-learning models for AFIB detection. These features include RR intervals and additional measures of irregularity and complexity.

\begin{itemize}
\item {\em RR Intervals} represent the time difference between two consecutive heartbeats. It is a fundamental feature in most AFIB detection research because RR interval variations indicate irregular heart rhythms. Analyzing the sequence of RR intervals helps identify the erratic patterns typical of AFIB \cite{sk2024deep}.

\item {\em GZIP Compression Length} transforms each bucket formed from the time series value into a UTF-8 character and concatenates them into a string of 41 characters. The length of the compressed string serves as a feature, providing a measure of the sequence's complexity and variability.

\item {\em Shannon Entropy (ShEn)} quantifies the unpredictability or randomness in the buckets formed from time series values. It is calculated as (\ref{eq-ShannEn}), where $n_i$ is the number of values in the $i$-th bucket, and $N$ is the total number of intervals. Higher entropy indicates more significant irregularity, which is characteristic of AFIB episodes.
\begin{equation}
\label{eq-ShannEn}
ShEn =  - \sum_{i=0}^{N-1}(\frac{n_i}{N}) * log(\frac{n_i}{N})    
\end{equation}

\item {\em Sample Entropy (SaEn)} is calculated using the TSFEL library processing the buckets formed from time series as input. Sample Entropy measures the complexity of time-series data, with higher values suggesting more irregular and complex patterns, typical of AFIB.
\end{itemize}

By leveraging these features, our approach aims to effectively capture the nuances of different heart rhythms, enhancing the model's ability to distinguish between AFIB and NonAFIB episodes. This comprehensive feature engineering strategy is pivotal in improving the accuracy and reliability of AFIB detection using machine learning techniques.

To train the model effectively, we use only clean segments \cite{dojchinovski2020segment} and apply a majority rule, where intervals with more than 51\% of heartbeats labeled as 1 (indicative of AFIB) are classified as AFIB.

%
\subsection{Ensemble Machine Learning Models}

Ensemble Machine Learning Models combine the predictions from multiple models to improve accuracy and robustness. In our earlier work \cite{9653409}, we analyzed several ensemble ML models, which were also employed in this study for AFIB detection. 

To ensure the robustness and effectiveness of these models, we employed a 10-fold cross-validation technique along with hyperparameter tuning.

\subsubsection{Decision Tree (DT)} creates a model based on simple decision rules inferred from the data features. It recursively splits the data into subsets based on feature values, maximizing the separation between classes at each node. Decision Trees are intuitive and easy to interpret, making them a good baseline for initial model development in AFIB detection \cite{bin2017detection}. The following hyperparameters were tuned in the optimization process: maximum depth in range (1,20), minimum samples per leaf (2,6), random state [42], criterion [gini, entropy], maximal features [sqrt,log2, None], and CCP alpha [0.0, 0.01 0.1, 0.2].

\subsubsection{Random Forest (RF)} is an ensemble method that builds a collection of Decision Trees, each trained on a randomly selected subset of the training data. By aggregating the predictions from multiple trees, Random Forest reduces overfitting and improves generalization. This method is particularly effective in dealing with noisy data and capturing complex heart rate variability patterns indicative of AFIB. \cite{kennedy2016automated}. The same hyperparameter settings were used as in DT, maximum depth in range (1,20), minimum samples per leaf (2,6), random state [42], criterion [gini,entropy], maximal features [sqrt,log2, None], and CCP alpha [0.0, 0.01 0.1, 0.2]

\subsubsection{XGBoost (XGB)} stands for Extreme Gradient Boosting as a powerful ensemble learning technique that builds decision trees in parallel, optimizing each tree based on the errors of the previous trees. XGBoost incorporates regularization techniques to prevent overfitting, making it highly effective for large datasets and complex feature spaces. Its ability to compute individual tree errors concurrently enhances model performance and generalization, which is crucial for accurately detecting AFIB episodes amidst varying heart rhythms  \cite{chen2018classification}. The hyperparameter tuning optimized the learning rate in the range [0.01, 0.1, 0.16, 0.2], number of estimators [100, 200, 300], max depth range(1,20), colsample bytree [0.5, 0.7 0.9], min child weight [1,2], and regularization parameter reg alpha [0.01, 0.1, 1].

\subsubsection{AdaBoost (AB)}  stands for Adaptive Boosting that combines multiple weak learners sequentially. It assigns higher weights to misclassified instances, forcing subsequent models to focus more on these challenging cases. This iterative process improves the model's ability to correct errors and enhances its robustness. It supports the classification of imbalanced datasets, such as those with fewer AFIB episodes compared to NonAFIB episodes, by focusing learning on the minority class \cite{mahmood2019comparison}. We optimized the number of estimators in the range [100, 200, 300] and learning rate [0.01, 0.1, 0.2, 0.3] within the hyperparameter tuning.

\subsubsection{CatBoost (CB)} means Categorical Boosting that handles categorical features effectively through an ordered boosting approach. It builds decision trees by considering the order of observations and incorporating categorical data directly into the model. CatBoost is known for its training efficiency and ability to handle high-cardinality categorical features, making it suitable for complex datasets encountered in AFIB detection. We selected the loss function [Logloss] and evaluation metric [AUC]. The hyperparameter tuning optimized the learning rate [0.01, 0.1, 0.2] and colsample by level [0.5, 0.7, 0.9].


\subsection{Deep Learning Model Development}

We also employed deep learning (DL) models to detect atrial fibrillation (AFIB) from ECG data. 

\subsubsection{Recurrent Neural Network (RNN)} are designed to process sequential data where the order of inputs matters, such as time series data. In AFIB detection, RNNs are advantageous because they can learn temporal dependencies and relationships between heartbeats, capturing the irregular patterns characteristic of AFIB episodes. RNNs maintain a hidden state updated at each time step, allowing them to retain information across long sequences \cite{sujadevi2018real}. They can effectively handle sequential data, such as ECG signals  \cite{andersen2019deep} because it is designed to maintain temporal dependencies between inputs. 

The RNN model used in this study consists of several layers. The SimpleRNN Layer is the initial layer in the network that processes input data as a sequence of time steps, capturing temporal patterns. The {\em dense layers} are created with Leaky ReLU Activations. After the SimpleRNN layer, the model incorporates six fully connected (Dense) layers. These layers use leaky ReLU (Rectified Linear Unit) activation functions to introduce non-linearity, allowing the network to learn more complex data patterns.
The {\em Dropout Layers} are interspersed between the dense layers to mitigate overfitting. These layers randomly deactivate some neurons during training, forcing the network to learn more robust features. {\em L2 Regularization} \cite{phaisangittisagul2016analysis} is applied to the dense layers to penalize large weights, further controlling overfitting by smoothing the model. Learning Rate Scheduler is a dynamic scheduler that adjusts the learning rate during training epochs to enhance model convergence and performance. {\em Optimizer and Loss Function} is realized by the Adam optimizer for efficient gradient-based optimization and binary cross-entropy loss function to measure the error in binary classification tasks (AFIB vs. NonAFIB). The {\em training parameters} include multiple epochs with a batch size of 128. The validation set, derived from a separate test set, assesses the model's generalization performance during training.

\subsubsection{Siamese Neural Network (SNN)} is based on a specialized architecture that learns to measure similarity or dissimilarity between two input instances, regardless of the specific task. In the context of AFIB detection, SNNs can be trained to compare segments of ECG data and determine whether they represent similar rhythm patterns (e.g., AFIB vs. NonAFIB) \cite{ng2023few}.

The {\em twin network architecture} in the SNN consists of two identical branches, each processing ECG signals. The architecture is designed such that both branches have the same structure and shared weights, ensuring that they learn the same features. Each branch separately consists of three fully connected Dense layers and four more merged layers. The {\em feature extraction and comparison} is done during training, where both branches receive pairs of ECG signals (one set from AFIB episodes and another from NSR episodes). The outputs from both branches are combined and used for the final classification.
Yje {\em loss Function and optimization} uses the binary cross-entropy loss function, suitable for binary classification tasks. The Adam optimizer updates the network weights during training, minimizing classification errors.
The {\em key advantage} is the dual-branch SNN architecture that allows the model to effectively learn temporal patterns and similarities between pairs of inputs, which enhances its ability to differentiate between AFIB and NSR episodes.

\subsubsection{Deep Neural Network (DNN)} is designed to learn and represent complex data patterns through multiple layers of neurons. It is particularly effective at extracting hierarchical features from raw input data, essential for making accurate predictions or classifications in AFIB detection. By leveraging deep architectures, DNNs can capture high-level abstractions and fine-grained details from ECG data, enhancing the model's ability to distinguish between different heart rhythms \cite{tran2020multifusionnet}.

We implement DNN architecture using Keras to extract and process features from the input data through multiple layers. 
The {\em dense layers with leaky ReLU activations} are crucial for capturing non-linear relationships in the input data and distinguishing between normal sinus rhythm and AFIB episodes. The {|em dropout layers} are applied to the dense layers to prevent overfitting. This regularization technique helps the model generalize unseen data better by randomly dropping units during training.
The {\em Optimizer and Loss Function} is realized by the Adam optimizer for training and minimizes the binary cross-entropy loss function. This setup is well-suited for binary classification tasks and helps the model learn effectively. The {\em training strategy} optimizes the model's ability to classify AFIB episodes accurately. 

\subsubsection{Hyperparemeter optimization} includes the following ranges: dropout rate [0.1 - 0.5], optimizer [adam,sgd], epochs [5-50], batch size [32 - 256], activation [relu, leaky relu] and layers config [3,9,12,18].

%

\subsection{Evaluation Methodology}

Evaluating the performance of machine learning models in detecting atrial fibrillation (AFIB) is critical, especially given the inherent class imbalance in AFIB datasets, where NonAFIB episodes often outnumber AFIB episodes. We use the F1 score to assess model performance, a metric particularly suited for imbalanced classification problems.

The \textbf{F1 score }is calculated as the harmonic mean of sensitivity (SEN) and precision, providing a balanced measure that considers both false positives and false negatives. This is especially important in medical diagnoses, where missing an AFIB episode (false negative) or incorrectly identifying a NonAFIB episode as AFIB (false positive) can have significant clinical implications. The F1 score is calculated by (\ref{eq-f1}), Where: \textbf{Sensitivity} (SEN), also known as the True Positive Rate (TPR), measures the proportion of actual positive cases (AFIB episodes) correctly identified by the model calculated by (\ref{eq-sen}), and \textbf{Precision}, also known as Positive Predictive Value (PPV), measures the proportion of predicted positive cases versus all positives, calculated by (\ref{eq-sen}).

\begin{equation}
\label{eq-f1}
F_1 = \frac{2 * PPV * SEN}{PPV + SEN}
\end{equation}
\begin{equation}
\label{eq-sen}
SEN = \frac{TP}{TP + FN} \qquad PPV = \frac{TP}{TP + FP}
\end{equation}

\subsubsection{Duration-Based Evaluation Method} \cite{gusev2019performance} evaluates the duration of correctly identified AFIB episodes instead of the heartbeat-oriented evaluation, providing a more nuanced understanding of the model's ability to detect AFIB over varying time intervals.

Each analyzed interval is assigned a value, and a specific post-processing procedure is applied to smooth out short sequences with alternating labels. This smoothing process helps mitigate the impact of noisy predictions and more accurately delineates the beginning and end of AFIB episodes. By refining these sequences, the model's detection of AFIB episodes aligns more closely with clinically meaningful patterns, providing a more reliable evaluation of its real-world applicability.

This dual approach ensures that the evaluation captures the model's ability to correctly identify AFIB cases and its effectiveness in recognizing AFIB episodes amidst varying heart rhythms.

\subsection{Experiments}

We specify three experiments based on different training/validation/ testing datasets.

\subsubsection{Experiment 1} uses a random selection of data samples and forms three subsets by an 80/10/10 split ratio. This splitting method lacks robustness and generality since it includes data from the same patient in all three datasets, so the model will be evaluated with data for the patient for which it was trained and validated. Experiment 1a uses MITDB and Experiment 1b LTAFDB.

\subsubsection{Experiment 2} uses different datasets where the first dataset forms the training and validation dataset with a ratio of 80/20 split and a different dataset for testing. We use MITDB as the first dataset divided into training and validation datasets. LTAFDB is used as a testing dataset. 
This method ensures testing with a completely independent dataset, providing a robust measure of its performance in real-world scenarios and generalization capability across different datasets and patient populations. Experiment 2a uses MITDB for training and LTAFDB for testing, while Experiment 2b uses the opposite LTAFDB for training and MITDB for testing.

\subsubsection{Experiment 3} uses the De Chazal Method, where the MITDB dataset is divided with the interpatient method into DS1 and DS2 datasets, specifically designed to handle the class imbalance present in the MITDB database, such that DS1 is used for training and DS2 is used for testing. The key aspect of this method is that patients included in the training dataset are not included in the testing set, preventing data leakage and overfitting.
This split method is highly effective in balancing patient distribution. The DS1 dataset includes patients 101, 106, 108, 109, 112, 114, 115, 116, 118, 119, 122, 124, 201, 203, 205, 207, 208, 209, 215, 220, 223, and 230, and DS2 the remaining from the MITDB.

%
%
\section{Results}
\label{sec-results}

Tables 1 through 5 present the results for each evaluated model across different experiments. 

\begin{table}[!h]
\centering
\caption{Results obtained from Experiment 1a, training with MITDB, and evaluating with MITDB}
\label{tab-results}%
\begin{tabular}{@{}lrcclrcc@{}}
\hline 
Method  & SEN (\%) & PPV (\%) & F1 (\%)  &  Mean (\%) & STD (\%) \\
\hline 
DT     &91.14	&87.71 &   89.39  & 73.02  & 32.66\\
\textbf{RF}     &91.66&    89.36 & \textbf{90.49}  & 88.52  & 1.14 \\
XGB    &91.35	&88.52 &   89.91  & 88.89  & 0.60  \\
AB     &93.10	&86.84 &   89.86  & 88.92  & 0.62 \\
CB     &91.56	&89.17 &   90.35  & 89.53  & 0.13 \\
RNN    &91.80	&87.24 &   89.46  & 89.46  & 0.00 \\
SNN    &97.44	&76.62 &   85.79  & 85.79  & 0.00 \\
DNN    &88.21	&89.03 &   88.62  & 88.62  & 0.00 \\
\hline 
\end{tabular}
\end{table}

Among all the models, the \textbf{Deep Neural Network (DNN)} model, as shown in Table ~\ref{tab-results3}, which was trained on the MITDB dataset and tested on the LTAFDB dataset, stands out as the most effective. It achieved an impressive F1 score of 95.18\% in detecting AFIB with separate training and testing datasets.

\begin{table}[!h]
\centering
\caption{Results obtained from Experiment 1b, training with LTAFDB, and evaluating with LTAFDB}
\label{tab-results}%
\begin{tabular}{@{}lrcclrcc@{}}
\hline 
Method  & SEN (\%) & PPV (\%) & F1 (\%)  &  Mean (\%) & STD (\%) \\
\hline 
DT     &96.56	&96.95    &96.75	&96.55	&0.13 \\
RF     &96.55	&97.05    &96.80	&96.68	&0.14 \\
XGB    &96.44	&97.11    &96.77	&96.74	&0.09 \\
AB     &96.82	&96.69    &96.75	&96.68	&0.06 \\
\textbf{CB}     &96.58	  &97.06    &\textbf{96.82}	&96.84	&0.01 \\
RNN    &96.29	&97.18    &96.73	&96.73	&0.00 \\
SNN    &96.61	&96.71    &96.66	&96.66	&0.00 \\
DNN    &96.44	&97.06    &96.75	&96.75	&0.00 \\
\hline 
\end{tabular}
\end{table}

%


\begin{table}[!h]
\centering
\caption{Results obtained from Experiment 2a, training with MITDB, and evaluating with LTAFDB}
\label{tab-results3}%
\begin{tabular}{@{}lrcclrcc@{}}
\hline 
Method  & SEN (\%) & PPV (\%) & F1 (\%)  &  Mean (\%) & STD (\%) \\
\hline 
DT     &90.90	&97.01   &93.85	  &72.99	&32.65\\
RF     &89.88	&97.25   &93.42	  &88.48	&1.09 \\
XGB    &89.52	&97.29   &93.24	  &88.78	&0.50 \\
AB     &91.78	&97.17   &94.40	  &88.85	&0.57 \\
CB     &89.58	&97.21   &93.24	  &89.49	&0.14 \\
RNN    &93.75	&96.48	 &95.09	  &95.09	&0.00 \\
SNN    &95.25	&94.58	 &94.91	  &94.91	&0.00 \\
\textbf{DNN}    &94.87	&95.49	 &\textbf{95.18}	  &95.18	&0.00 \\
\hline 
\end{tabular}
\end{table}

\begin{table}[!h]
\centering
\caption{Results obtained from Experiment 2b, training with LTAFDB, and evaluating with MITDB}
\label{tab-results}%
\begin{tabular}{@{}lrcclrcc@{}}
\hline 
Method  & SEN (\%) & PPV (\%) & F1 (\%)  &  Mean (\%) & STD (\%) \\
\hline 
DT     &96.43	&58.60    &72.89	&96.52	&0.13 \\
RF     &96.17	&59.18    &73.27	&96.65	&0.14 \\
XGB    &96.43	&58.92    &73.14	&96.73	&0.10 \\
AB     &96.92	&57.46    &72.15	&96.65	&0.06 \\
CB     &96.23	&58.71    &72.93	&96.82	&0.01 \\
RNN    &97.05	&58.00    &72.61	&72.61	&0.00 \\
\textbf{SNN}    &93.78	&64.36    &\textbf{76.34}	&76.34	&0.00 \\
DNN    &97.69	&56.00    &71.18	&71.18	&0.00 \\
\hline 
\end{tabular}
\end{table}

\begin{table}[!h]
\centering
\caption{Results obtained from Experiment 3, training with DS1, and evaluating with DS2}
\label{tab-results}%
\begin{tabular}{@{}lrcclrcc@{}}
\hline 
Method  & SEN (\%) & PPV (\%) & F1 (\%)  &  Mean (\%) & STD (\%) \\
\hline 
DT     &87.35	&79.93    &83.47	&53.10	&37.58 \\
\textbf{RF}     &89.20	&79.74    &\textbf{84.21}	&59.59	&35.96 \\
XGB    &83.24	&81.20    &82.21	&80.62	&1.88 \\
AB     &81.29	&82.29    &81.79	&82.22	&0.97 \\
CB     &79.67	&82.23    &80.93	&80.51	&0.59 \\
RNN    &91.04	&75.30    &82.43	&82.43	&0.00 \\
SNN    &92.14	&68.87    &78.82	&78.82	&0.00 \\
DNN    &88.08	&73.34    &80.04	&80.04	&0.00 \\
\hline 
\end{tabular}
\end{table}

The \textbf{DNN} is the best-performing model due to its exceptional ability to generalize across different datasets. This model performed well on distinct datasets and demonstrated robustness by including cross-validation and an additional validation dataset in the evaluation process.

%
%

\section{Discussion}
\label{sec-discussion}

Based on the results obtained from our experiments and the reviewed studies, NNs have proven effective in detecting AFIB. 

To ensure robustness, we used different variations of training and testing of the models to prove their efficiency and accuracy. Due to the imbalance in our datasets, we selected the F1 score as our performance metric, as it is well-suited for assessing model performance on skewed datasets. We also used 10-fold cross-validation on our ensembling methods, mean and standard deviation of the classifier performance,  and additional hyperparameter tuning on these methods.

The success of our approach also hinges on selecting appropriate parameters and designing an optimal NN structure tailored for AFIB detection. 
While the SNN, RNN, and DNN methods also performed well, the CB model demonstrated superior performance in our study, with an F1 score of 96.82\%.
Table~\ref{tab-comparison} compares the results from related research achieved by other authors. Note that our methods did not achieve the highest scores compared to the others. Still, we aimed to attain maximal robustness using a 10-fold cross-validation and addition validation dataset. We also included hyperparameter tuning to reach the best results from our models. Additionally, we applied extra techniques to prevent overfitting.

\begin{table}[!t]
\caption{Comparison with other papers}
\label{tab-comparison}%
\begin{tabular}{@{}lcrcl@{}}
\hline 
Related Work & Method  & F1(\%) & Split & Dataset \\
\hline 
Martis et al.\cite{martis2013application}    & ANN   & 98.97  & - & MITDB, AFDB \\
Wang et al.\cite{wang2020atrial}  & CNN+LSTM   & 82.00  & 70/30 & MITDB   \\
Teijeiro et al.\cite{teijeiro2017arrhythmia}    & RNN  & 83.00 & - & PhysioNet/CinC 2017   \\
Cai et al.\cite{cai2020accurate}  & DDNN  & 90.70 & 5 fold CV & proprietary data   \\
Faust et al.\cite{faust2018automated}    & RNN + LSTM  & 98.35 & 10 fold CV & AFDB   \\
Artis et al.\cite{artis1991detection}    & ANN  & 92.60 & - & MITDB   \\
\textbf{Our method}    & RF  & 90.49 & 80/10/10+10 fold CV & MITDB   \\
\textbf{Our method}    & CB  & 96.82 & 80/10/10+10 fold CV & LTAFDB  \\
\textbf{Our method}    & DNN  & 95.18 & separate datasets & MITDB, LTAFDB   \\
\textbf{Our method}    & SNN  & 76.34 & separate datasets & LTAFDB, MITDB   \\
\textbf{Our method}    & RF  & 84.21 &  deChazal split & MITDB   \\
\hline 
\end{tabular}
\end{table}

Table~\ref{tab-optimal} represents the optimal hyperparameters used to achieve the highest result for each train/test combination we used, as shown in Table ~\ref{tab-comparison}.

\begin{table}[!t]
\centering
\caption{Optimal hyperparameters for the results achieved in Table \ref{tab-comparison}}
\label{tab-optimal}
\begin{tabular}{@{}lccp{8cm}@{}}
\hline 
Model & Training & Testing & Optimal Parameters \\
\hline 
RF & MITDB & MITDB & \begin{tabular}[t]{@{}l@{}}ccp alpha:[0.0], criterion:[gini], max depth:[12], \\ max features:[sqrt], estimators:[300], random state [42]\end{tabular} \\
CB & LTAFDB & LTAFDB & \begin{tabular}[t]{@{}l@{}}colsample bylevel:[0.5], evaluation metric:[AUC], \\ learning rate:[0.1], loss function:[Logloss] \end{tabular} \\
DNN & MITDB & LTAFDB & \begin{tabular}[t]{@{}l@{}}layers\_config=[3, 9, 12, 18],  dropout rate:[0.1], \\ optimizer:[adam], epochs:[19], batch size:[128]\end{tabular} \\
SNN & LTAFDB & MITDB & \begin{tabular}[t]{@{}l@{}}layers\_config=[3, 9, 12, 18], epochs:[8], batch size:[128], \\ optimizer:[adam]\end{tabular} \\
RF & DS1 & DS2 & \begin{tabular}[t]{@{}l@{}}ccp alpha:[0.0], criterion:[entropy], max depth:[2], \\ max features:[sqrt], estimators:[200], random state:[42]\end{tabular} \\
\hline 
\end{tabular}
\end{table}

\section{Conclusion}
\label{sec-conclusion}

We presented a new method for developing ML models based on newly derived features, such as Gzip length, Shannon Entropy, and Spectral Entropy, besides the series of differences in heart rates between consecutive heartbeats. 
Special labeling techniques divide the dataset into five classes, and the AFIB binary classification determines the start and end of an AFIB rhythm applied to the duration-based evaluation. We used the MITDB and LTAFDB datasets for the experiments and tried all the combinations to see which gave us the best results.
Specifically, the decision to train the model only with clean intervals improved the model's performance.
The results outperform other research compared with the same evaluation methodology. 

This research shows that DNN models have great potential in biomedical signal processing, especially for detecting AFIB. Advanced deep learning techniques are promising technology after carefully designing the underlying problem. These advancements offer hope for improving medical outcomes and patient care in heart health. Refining NN methods further could lead to even more progress in AFib detection and other vital areas of medical research. 
However, besides the secret ingredients in developing the model, we conclude that feature engineering improved the whole process, especially selecting the training subset from the training dataset by applying a specific labeling method.

\balance
\bibliographystyle{unsrt}  
\bibliography{references}

\end{document}